\pdfoutput=1

\documentclass[11pt]{article}

\usepackage{naacl2021}

\usepackage{times}
\usepackage{latexsym}

\usepackage[T1]{fontenc}

\usepackage[utf8]{inputenc}

\usepackage{microtype}

\usepackage{color}
\usepackage{amsmath,amsfonts}
\usepackage[export]{adjustbox}
\usepackage{subfig}
\usepackage{mathtools}
\usepackage{url}
\usepackage{float}
\usepackage{placeins}

\usepackage{stfloats}

\title{How do lexical semantics affect translation? An empirical study}

\author{Vivek Subramanian \\
  Amazon Alexa AI \\
  \texttt{viveksub@amazon.com} \\\And
  Dhanasekar Sundararaman \\
  Duke University \\
  \texttt{ds448@duke.edu} \\}

\begin{document}
\maketitle
\begin{abstract}
Neural machine translation (NMT) systems aim to map text from one language into another. While there are a wide variety of applications of NMT, one of the most important is translation of \textit{natural language}. A distinguishing factor of natural language is that words are typically ordered according to the rules of the grammar of a given language. Although many advances have been made in developing NMT systems for translating natural language, little research has been done on understanding how the word ordering of and lexical similarity between the source and target language affect translation performance. Here, we investigate these relationships on a variety of low-resource language pairs from the OpenSubtitles2016 database, where the source language is English, and find that the more similar the target language is to English, the greater the translation performance. In addition, we study the impact of providing NMT models with part of speech of words (POS) in the English sequence and find that, for Transformer-based models, the more dissimilar the target language is from English, the greater the benefit provided by POS.
\end{abstract}

\section{Introduction}
Neural machine translation (NMT) systems map text from one language into another via a neural network. Several approaches to NMT have been developed, commonly consisting of an encoder-decoder architecture and an attention mechanism \cite{sutskever2014sequence,bahdanau2014neural,luong2015effective}. The encoder seeks to extract a representation for the source sequence that captures all relevant semantics in the sequence. The decoder then utilizes this representation to generate a sequence of words, which is the translation. Attention allows the decoder to weight individual tokens in the source sequence depending on their importance to the word being generated. The Transformer \cite{vaswani2017attention}, a more complex architecture which employs multi-headed self- and cross-attention\footnote{Among other features, including positional encoding and layer normalization}, leading to a new state-of-the-art, has led to an unprecedented wave of research in NMT.

While these systems have shown a great deal of promise, relatively little has been done to understand in detail how the \textit{lexical semantics} of natural language, including \textit{lexical similarity} and \textit{word order} (\textit{e.g.}, subject-verb-object, or SVO), of the source and target language affects translation performance. For instance, \cite{johnson2017google} apply Google NMT, an LSTM-based architecture, to multilingual translation, achieving zero-shot learning between related languages Portuguese $\leftrightarrow$ Spanish and Korean $\leftrightarrow$ Japanese. However, little discussion is provided as to how the syntactic features of these languages contribute to performance. Recently, \cite{aharoni2019massively} utilized the Transformer architecture for multilingual translation on an in-house dataset consisting of 58 low-resource language pairs. While they acknowledge that the diversity of linguistic features can induce a bottleneck due to limited model capacity, they do not make any effort to analyze similarity between the source and target languages or to prune the languages on which their multilingual models are trained.

Hence, in this paper, we provide two main contributions. \textbf{First}: we perform an empirical study of translation from English into 15 languages of various word orders and degrees of lexical similarity to English. We utilize both LSTM- and Transformer-based models for our study. \textbf{Second}: we compare and contrast how explicit supervision with part of speech (POS), a grammatical feature closely tied to word ordering, of source-side tokens affects translation performance of these respective models. With insight into these trends, we hope to further research on multilingual translation by highlighting the importance of accounting directly for differences in lexical semantics during model development.






\section{Methods}

\subsection{Models}
For our experiments, we utilize LSTM- and Transformer-based architectures and refer the reader to \cite{sutskever2014sequence} and \cite{vaswani2017attention} for details. For each parallel corpus, words are first tokenized into subwords with byte pair encoding \cite{sennrich2015neural} using a common vocabulary for source and target languages. The LSTM and Transformer baselines allow us to study general trends in performance as a function of word order and lexical similarity. We then introduce two related models denoted LSTM\textsubscript{POS} and Transformer\textsubscript{POS} to which we directly provide part of speech (POS) of each word in the source sequence, obtained using \texttt{spaCy}\footnote{https://spacy.io/}. As POS is closely tied to word order, we hypothesize that providing POS of the source sequence to the decoder will improve alignment, attention, and ultimately, text generation, especially for languages whose word order is substantially different from the source. We assign the POS of each word to its subword tokens and append trainable POS embedding vectors to the subword token embeddings in a manner similar to \cite{sennrich2016linguistic,sundararaman2019syntax}.

\begin{table*}[t]
\begin{tabular}{c|c|c|c|c|c|c}
\begin{tabular}[c]{@{}c@{}}Target\\ Language\end{tabular} & \begin{tabular}[c]{@{}c@{}}Word\\ Order\end{tabular} & \begin{tabular}[c]{@{}c@{}}Levenshtein\\ distance\end{tabular} & \begin{tabular}[c]{@{}c@{}}Num.\\ train\end{tabular} & \begin{tabular}[c]{@{}c@{}}Num.\\ test\end{tabular} & \begin{tabular}[c]{@{}c@{}}LSTM:\\ BLEU score\end{tabular} & \begin{tabular}[c]{@{}c@{}}Transformer:\\ BLEU score\end{tabular} \\ \hline
Sinhala   (SI) & SOV & 0.642 & 540990 & 6075 & 11.36 & 12.76 \\ \hline
Bengali   (BN) & SOV & 0.632 & 372240 & 4138 & 11.60 & 13.41 \\ \hline
Hindi   (HI) & SOV & 0.632 & 83700 & 946 & 22.26 & 23.80 \\ \hline
Malayalam   (ML) & SOV / Flexible & 0.708 & 348120 & 3936 & 7.35 & 8.12 \\ \hline
Korean   (KO) & SOV / Flexible & 0.468 & 1251990 & 14001 & 5.66 & 6.85 \\ \hline
Basque   (EU) & Flexible / SOV & 0.407 & 725130 & 8137 & 14.98 & 16.60 \\ \hline
Georgian   (KA) & Flexible & 0.621 & 177910 & 2077 & 10.57 & 11.79 \\ \hline
Chinese   (ZH\_CN) & Flexible & 0.519 & 450000 & 5000 & 6.84 & 7.73 \\ \hline
Esperanto   (EO) & Flexible / SVO & 0.398 & 57960 & 729 & 11.31 & 12.65 \\ \hline
Latvian   (LV) & SVO / Flexible & 0.416 & 467550 & 5248 & 17.63 & 19.71 \\ \hline
Galician   (GL) & SVO / Flexible & 0.390 & 183150 & 2085 & 15.84 & 17.86 \\ \hline
Ukrainian   (UK) & SVO & 0.539 & 789930 & 8857 & 11.46 & 13.61 \\ \hline
French   (FR) & SVO & 0.386 & 450000 & 5000 & 22.53 & 23.95 \\ \hline
German   (DE) & SVO & 0.383 & 450000 & 5000 & 19.43 & 20.57 \\ \hline
Catalan   (CA) & SVO & 0.382 & 434250 & 4923 & 27.30 & 29.23
\end{tabular}
\caption{Target languages selected from OpenSubtitles2016 database, sorted first by word order then inversely by normalized Levenshtein distance to English, whose word ordering is SVO. Some languages have more than one dominant word order and are listed as such, with the most dominant appearing first. For each language, the number of training and testing examples is listed, along with the BLEU score achieved by the baseline LSTM and Transformer models.}
\label{tab:languages}
\end{table*}

\subsection{Dataset}

We translate from English into 15 target languages from the OpenSubtitles2016 database, which consists of human annotated captions for movies and films \cite{lison2016opensubtitles2016}. Unlike other publicly available parallel corpora such as Europarl \cite{koehn2005europarl}, JRC-Acquis \cite{steinberger2006jrc}, or WIT\textsuperscript{3} \cite{cettolo2012wit3}, the OpenSubtitles2016 database contains data from languages with a more diverse range of word orderings and lexical similarity, allowing us to robustly test our hypothesis. Of the 15 target languages chosen (see Table \ref{tab:languages}), 12 (Sinhala, Bengali, Hindi, Malayalam, Korean, Basque, Georgian, Esperanto, Latvian, Galician, Ukrainian, and Catalan) are considered low-resource since they have roughly 1 million sentence pairs or less, and 3 (Simplified Chinese, French, and German) are considered high-resource. We choose to focus primarily on low-resource languages as the performance of NMT systems has already achieved near human-level performance in data-rich settings \cite{wu2016google,hassan2018achieving}. Thus, low-resource language pairs stand to benefit the most from novel improvements to NMT architectures that harness specific features of the source and target languages. The remaining 3 languages were subsampled to 500K datapoints for consistency.

\subsection{Word order and lexical similarity}
We compare translation performance between our baselines and POS-augmented models as a function of two key features: \textit{word order} and \textit{lexical similarity}. For each target language, word order was obtained using Glottolog \cite{hammarstrom2017glottolog}, a professionally curated online catalog of the world's languages.\footnote{See Appendix \ref{app:refs} for a list of references pertinent to each target language.} These are listed in Table \ref{tab:languages}. Target languages fall into three major categories: subject-object-verb (SOV), flexible, or subject-verb-object (SVO). For languages that are flexible in structure but which still possess a prevalent form, both ``flexible'' and the prevalent form are listed, with whichever is more dominant appearing first. This allows us to view the languages on a spectrum, with SOV languages on one end and SVO languages on the other -- closest to English, whose word order is SVO.

We compute lexical similarity between source and target documents via normalized Levenshtein distance \cite{jan2007receptive}: the Levenshtein distance first computes the minimum number of character modifications (insertion, replacement, or deletions) that must be performed to map from the source to the target, and this value is then normalized by the average of the number of characters in the source and target. Levenshtein distance can be computed exactly in $\mathcal{O}(nm)$ time, where $n$ is the number of characters in the source and $m$ is the number of characters in the target. This metric makes the implicit assumptions that two languages are lexically similar when (1) their vocabularies consist of similar phonemes and (2) the phonemes are composed of similar numbers of characters. This simplified definition affords us the ability to compare languages whose alphabets may have no common characters without having to resort to manual transliteration of symbols into a common space of phonemes.

\subsection{Training details}
We utilize the OpenNMT-py LSTM and Transformer implementations \cite{klein-etal-2017-opennmt} and specify identical training conditions for each model type when training models on all 15 corpora. Specifically, for the LSTM models, we train using a Titan XP GPU for 75,000 steps using an embedding dimension of 512, hidden state dimension of 512, batch size of 64, and dropout rate of 0.1. For the Transformer models, we train for 75,000 steps using the same embedding and hidden state dimensions, batch size of 4096, and same dropout rate. For both models, POS embeddings were merged with subword token embeddings using a feature vector exponent of 0.7, resulting in roughly 5 to 8 dimensions being allocated for POS, depending on the corpus. We train all models with an Adam optimizer with $\eta = 1$, $\beta_1 = 0.9$, $\beta_2 = 0.999$, and $\epsilon = 10^{-7}$.

\begin{figure*}[t]
    \centering
    \subfloat[LSTM]{
        \centering
        \includegraphics[width=0.49\textwidth,valign=c]{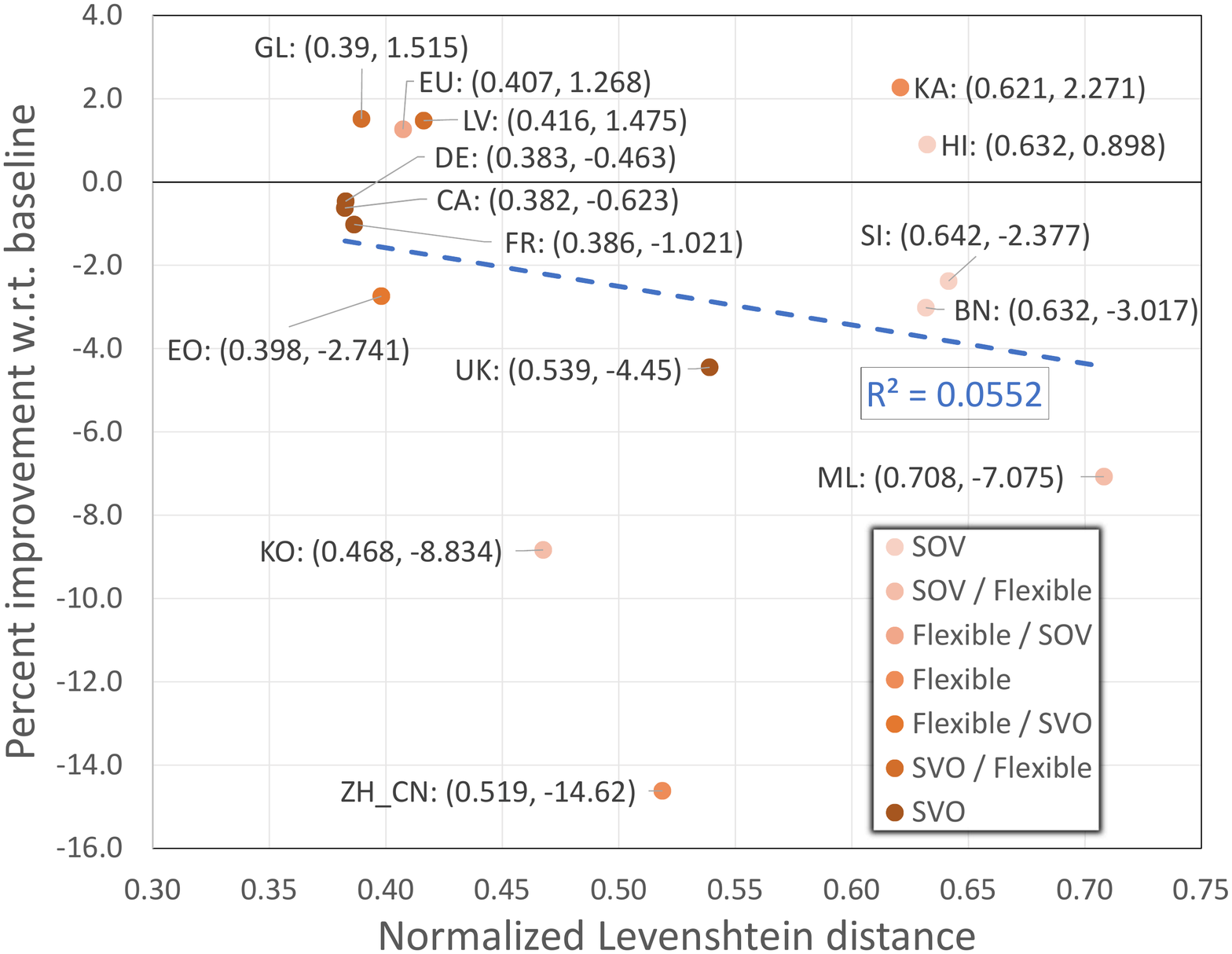}
        \label{fig:perfvsjsd}}
    \subfloat[Transformer]{
        \centering
        \includegraphics[width=0.49\textwidth,valign=c]{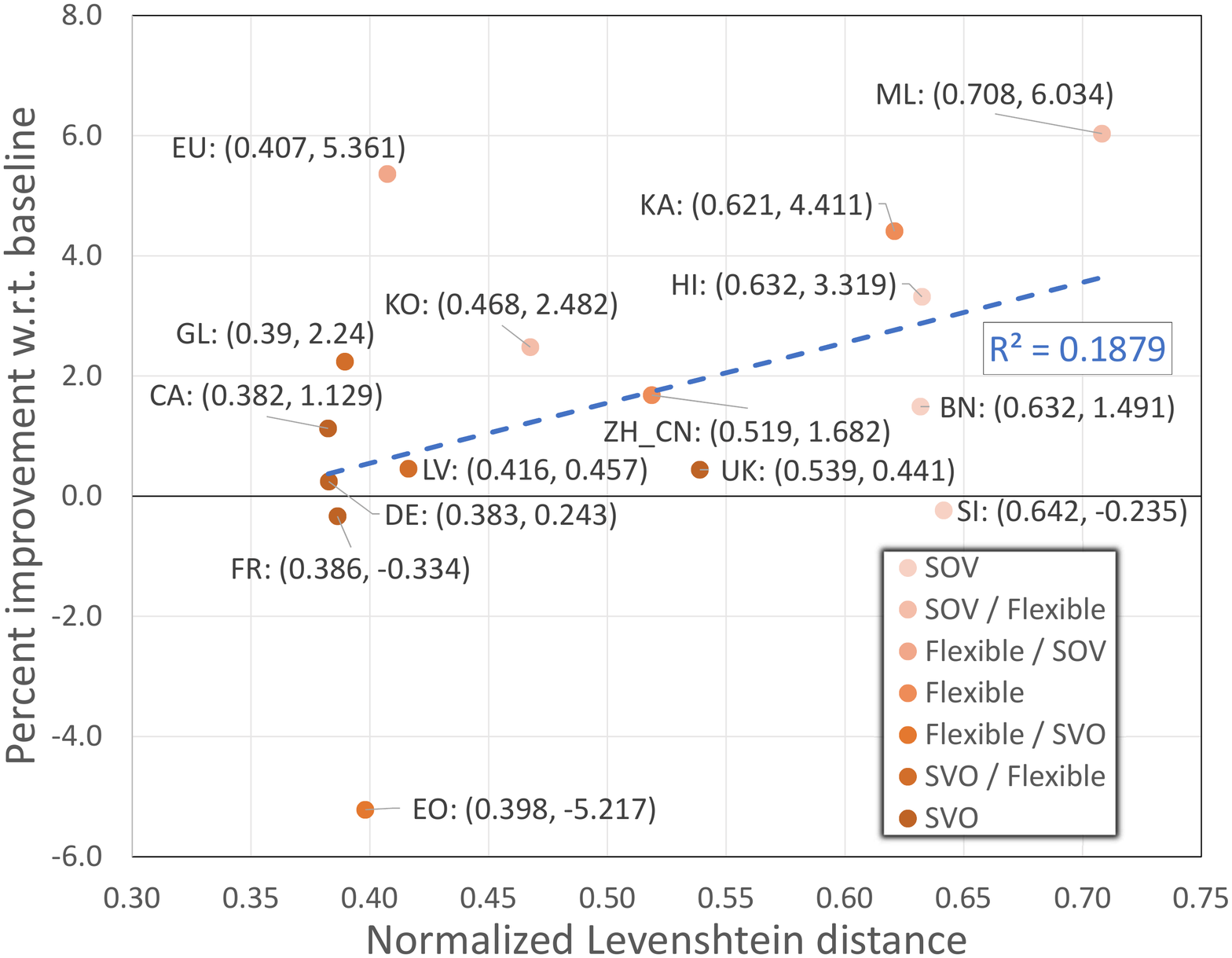}
        \label{fig:spansizevsjsd}}
    \caption{Performance improvements with POS models. Baseline BLEU scores given in Table \ref{tab:languages}. Shades of red indicate word order of target language. Dashed blue line-of-best-fit indicates overall correlation.}
    \label{fig:overallperformance}
\end{figure*}

\section{Results and Discussion}
We evaluate translation performance with BLEU scores \cite{papineni2002bleu} and report baseline results in Table \ref{tab:languages} for each of the fifteen target languages. As expected, we find that translation performance of both models drops as the word order of the target language differs more from that of English, which is SVO (Mann-Whitney-Wilcoxon (MWW) test: $p = 0.07$ for LSTM, $p = 0.05$ for Transformer; see Appendix \ref{app:stats} for details). We also find that translation performance decays roughly linearly with Levenshtein distance (Pearson's $r = -0.47$, $p = 0.08$ for both models).

HI and UK are outliers within their respective word order groups. The reduced performance on EN$\rightarrow$UK can be explained by the stark difference in lexical similarity of UK to EN compared to FR, DE, and CA. For EN$\rightarrow$HI, regressions of BLEU score against (1) number of testing samples and (2) number of unique English words in each of the 15 parallel corpora found no significant trends, ruling out biases (1) in the size of the test set or (2) due to overall vocabulary size. Thus, we believe the difference originates from unique aspects of HI that set it apart from SI and BN. For instance, conjugation of verbs in BN is much more subtle, often requiring changes of just a single syllable; in contrast, HI conjugations often require the addition of an extra word. In addition, in HI, both determiners and verbs are gendered (\textit{e.g.}, ``he eats the apple'' is different from ``she eats the orange''). The greater number of distinguishing factors reduce the overall entropy of predicting words in HI, leading to the increase in BLEU score.

Figure 1 depicts the change in performance of each of these architecture styles when POS is included as an input feature. Notably, the LSTM\textsubscript{POS} models generally perform worse compared to their baseline counterparts (mean difference of -2.51\%; $t$-test, $p=0.06$) while the Transformer\textsubscript{POS} models perform significantly better (mean difference of +1.64\%; $t$-test, $p=0.02$). Furthermore, the degradation of the LSTM\textsubscript{POS} performance worsens slightly as the word order of the target language becomes more different (MWW test, $p=0.15$) and as lexical similarity decreases (Pearson's $r = -0.24$, $p=0.40$). On the other hand, gains seen by Transformer\textsubscript{POS} improve more and more as the disparity between source and target languages increases (MWW test, $p=0.02$ for word order; Pearson's $r=+0.43$, $p=0.11$ for lexical similarity). Thus, while the LSTM may be able to infer semantic relationships between distant tokens that are preserved during decoding, its reduced model capacity relative to the Transformer renders it inept at generating syntactically correct sequences as the ordering of target words begins to change. In contrast, the Transformer utilizes source POS features as anchor points to effectively learn the word ordering of source sequence (self-attention) and to perform better alignment during decoding (cross-attention).

\section{Conclusion}
In conclusion, we have demonstrated that both LSTMs and Transformers perform best at NMT of natural language when the source and target languages possess similar lexical semantics. In addition, incorporating POS as an input feature to the Transformer helps the model align words, especially when the source and target word orderings are severely mismatched. Future work will be towards incorporating these findings into multilingual translation models of low-resource language pairs. One of the most common approaches for this objective is bridging, in which translation between two disparate languages $A$ and $C$ with little or no parallel training data is accomplished by introducing one or more intermediary languages $B_i$ with sufficient parallel training data. Our results can be used to inform which language pairs are likely to be helpful if selected as intermediaries, based on similarities in lexical semantics. For many of these languages, POS parsers may not be readily available. Hence, we are also investigating transfer / multi-task learning approaches to sharing available POS information across several languages.

\bibliographystyle{acl_natbib}
\bibliography{naacl2021}

\clearpage
\newpage
\appendix
\newpage
\section{References for individual language word orderings}
\label{app:refs}

In Table \ref{tab:glottologrefs}, we provide a list of references for each of the 15 target languages. Most were obtained using Glottolog \cite{hammarstrom2017glottolog}, with the exceptions of Galician and Ukrainian. Each language is classified as subject-object-verb (SOV), subject-verb-object (SVO), or flexible. For languages that are considered flexible but have one more prevalent word ordering, both are listed, with the dominant one appearing first. In addition, the source language English is SVO \cite{huddleston2002cambridge}.

\begin{table*}[!b]
\centering
\begin{tabular}{c|c|c}
Target   language & Word order & Reference \\ \hline
Sinhala   (SI) & SOV & \cite{reynolds1980sinhalese} \\ \hline
Bengali   (BN) & SOV & \cite{bhattacharja2007word} \\ \hline
Hindi   (HI) & SOV & \cite{koul2008modern} \\ \hline
Malayalam   (ML) & SOV / Flexible & \cite{syamala1981intensive} \\ \hline
Korean   (KO) & SOV / Flexible & \cite{sohn2001korean} \\ \hline
Basque   (EU) & Flexible / SOV & \cite{hualde2011grammar} \\ \hline
Georgian   (KA) & Flexible & \cite{hewitt1995georgian} \\ \hline
Chinese   (ZH\_CN) & Flexible & \cite{li1989mandarin} \\ \hline
Esperanto   (EO) & Flexible / SVO & \cite{gledhill1998grammar} \\ \hline
Latvian   (LV) & SVO / Flexible & \cite{mathiassen1997short} \\ \hline
Galician   (GL) & SVO / Flexible & \cite{ud2020ud} \\ \hline
Ukrainian   (UK) & SVO & \cite{jenkala2007read} \\ \hline
French   (FR) & SVO & \cite{resnick2012essential} \\ \hline
German   (DE) & SVO & \cite{curme1905grammar} \\ \hline
Catalan   (CA) & SVO & \cite{forcadell2013subject}
\end{tabular}
\caption{References used to obtain word order for each language.}
\label{tab:glottologrefs}
\end{table*}

\section{Mann-Whitney-Wilcoxon test for ordinality}
\label{app:stats}

To evaluate whether samples from each word ordering set were significantly greater or less than those from another, we employed the Mann-Whitney-Wilcoxon (MWW) test. We first grouped together all languages that were at least predominantly SOV into a single group (Group 1); these included SI, BN, HI, ML, KO, and EU. We next grouped the languages classified as having a flexible word order (Group 2); these included KA and ZH\_CN. Finally, we grouped all languages that were at least predominantly SVO (Group 3); these included EO, LV, GL, UK, FR, DE, and CA.

We ran three pairwise MWW tests between Groups 1 and 2, Groups 2 and 3, and Groups 1 and 3. These were one-sided test, with the left tail chosen when we wanted to test whether samples from the first group were less than samples from the second, and vice versa. $p$-values are reported for all tests in Tables \ref{tab:utestrawperf}-\ref{tab:utestimrpov2} below. In each case $g_1$, $g_2$, and $g_3$ correspond to samples from Groups 1, 2, and 3, respectively.

\begin{table}[H]
\centering
\begin{tabular}{c|c|c}
 & LSTM & Transformer\\ \hline
$P(g_1 < g_2)$ & 0.857 & 0.086\\ \hline
$P(g_2 < g_3)$ & 0.028 & 0.028\\ \hline
$P(g_1 < g_3)$ & 0.069 & 0.051
\end{tabular}
\caption{$p$-values for MWW test for ordinality of baseline model performance as a function of word order.}
\label{tab:utestrawperf}
\end{table}

\begin{table}[H]
\centering
\begin{tabular}{c|c}
 & LSTM \\ \hline
$P(g_1 < g_2)$ & 0.057 \\ \hline
$P(g_2 < g_3)$ & 0.556 \\ \hline
$P(g_1 < g_3)$ & 0.147
\end{tabular}
\caption{$p$-values for MWW test for ordinality of performance improvement of LSTM\textsubscript{POS} model over baseline as a function of word order.}
\label{tab:utestimrpov1}
\end{table}

\begin{table}[H]
\centering
\begin{tabular}{c|c}
 & Transformer \\ \hline
$P(g_1 > g_2)$ & 0.571 \\ \hline
$P(g_2 > g_3)$ & 0.056 \\ \hline
$P(g_1 > g_3)$ & 0.017
\end{tabular}
\caption{$p$-values for MWW test for ordinality of performance improvement of Transformer\textsubscript{POS} model over baseline as a function of word order.}
\label{tab:utestimrpov2}
\end{table}

\end{document}